\documentclass[letterpaper, 10 pt, conference]{ieeeconf}  

\IEEEoverridecommandlockouts                              

\usepackage{graphics} 
\usepackage{graphicx} 
\usepackage{mathptmx} 
\usepackage{cite}
\usepackage{amsmath} 
\usepackage{amssymb}  
\usepackage{bm}
\usepackage{url}
\usepackage{tabularx}
\usepackage{array}

\title{\LARGE \bf Design and Development of a Remotely Wire-Driven Walking Robot}

\author{Takahiro Hattori$^{1}$, Kento Kawaharazuka$^{1,2}$, Kei Okada$^{1}$
\thanks{*This work was not supported by any organization}
\thanks{$^{1}$The authors are with the Department of Mechano-Informatics, Graduate School of Information Science and Technology, The University of Tokyo, 7-3-1 Hongo, Bunkyo-ku, Tokyo, 113-8656, Japan.{\tt\small [t-hattori,
kawaharazuka, k-okada] @ jsk.imi.i.u-tokyo.ac.jp}}%
\thanks{$^{2}$The author is with the AI Center, Graduate School of Information Science and Technology, The University of Tokyo}%
}

\begin{document}
\newcommand{\TODO}[1]{\fbox{{\textbf TODO:}#1}}
\newcommand{\secref}[1]{Section \ref{#1}}
\newcommand{\subsecref}[1]{Subsection \ref{#1}}
\newcommand{\tabref}[1]{{Table \ref{#1}}}
\newcommand{\figref}[1]{{Fig. \ref{#1}}}
\newcommand{\equref}[1]{{Eq. \ref{#1}}}
\newcommand{\algoref}[1]{{Alg. \ref{#1}}}
\newcommand{\enumref}[1]{{\ref{#1}}}

\maketitle
\thispagestyle{empty}
\pagestyle{empty}

\begin{abstract}
Operating in environments too harsh or inaccessible for humans is one of the critical roles expected of robots. However, such environments often pose risks to electronic components as well. To overcome this, various approaches have been developed, including autonomous mobile robots without electronics, hydraulic remotely actuated mobile robots, and long-reach robot arms driven by wires. Among these, electronics-free autonomous robots cannot make complex decisions, while hydraulically actuated mobile robots and wire-driven robot arms are used in harsh environments such as nuclear power plants. Mobile robots offer greater reach and obstacle avoidance than robot arms, and wire mechanisms offer broader environmental applicability than hydraulics. However, wire-driven systems have not been used for remote actuation of mobile robots. In this study, we propose a novel mechanism called Remote Wire Drive that enables remote actuation of mobile robots via wires. This mechanism is a series connection of decoupled joints, a mechanism used in wire-driven robot arms, adapted for power transmission. We experimentally validated its feasibility by actuating a wire-driven quadruped robot, which we also developed in this study, through Remote Wire Drive.
\end{abstract}


\section{Introduction}

In recent years, motors and sensors such as encoders have been used in most robots. However, these electronic components are vulnerable to special environments such as high humidity, radiation, heat, high voltage, or gas-filled spaces \cite{hazard_protection} \cite{electronics_harsh}. Since robots are expected to operate in such harsh environments where humans cannot enter, technologies that overcome these weaknesses are essential.

One such approach is the development of robots that operate without electronics. For example, the robot developed by Drotman et al. \cite{electronics_free_air_powered} uses pneumatic circuits to move three pneumatic legs in specific patterns, enabling walking and direction switching based on environmental contact. Similarly, He et al. \cite{modular_electfree} achieved electronics-free autonomous control by combining origami structures with materials that respond and deform in reaction to light, heat, or chemical stimuli. However, these electronics-free information processing systems are significantly larger in mass and volume than their electronic counterparts and can only perform simple behaviors.

An alternative approach involves remotely actuated mobile robots. For instance, the nuclear power plant robot HUMALT developed by Nakamura et al. \cite{HUMALT_RSJ} is a hydraulically driven mobile robot without onboard electronics. It is controlled remotely through long hydraulic tubes from a control unit and pump located tens of meters away. This remote actuation method allows both the elimination of electronics from harsh environments and the retention of complex computer control. However, hydraulic systems also present challenges such as freezing below 0°C, boiling above 100°C, the requirement for watertight components, and the mechanical and systemic complexity of pumps and valves.

\begin{figure}[t]
  \centering
  \includegraphics[width=\linewidth]{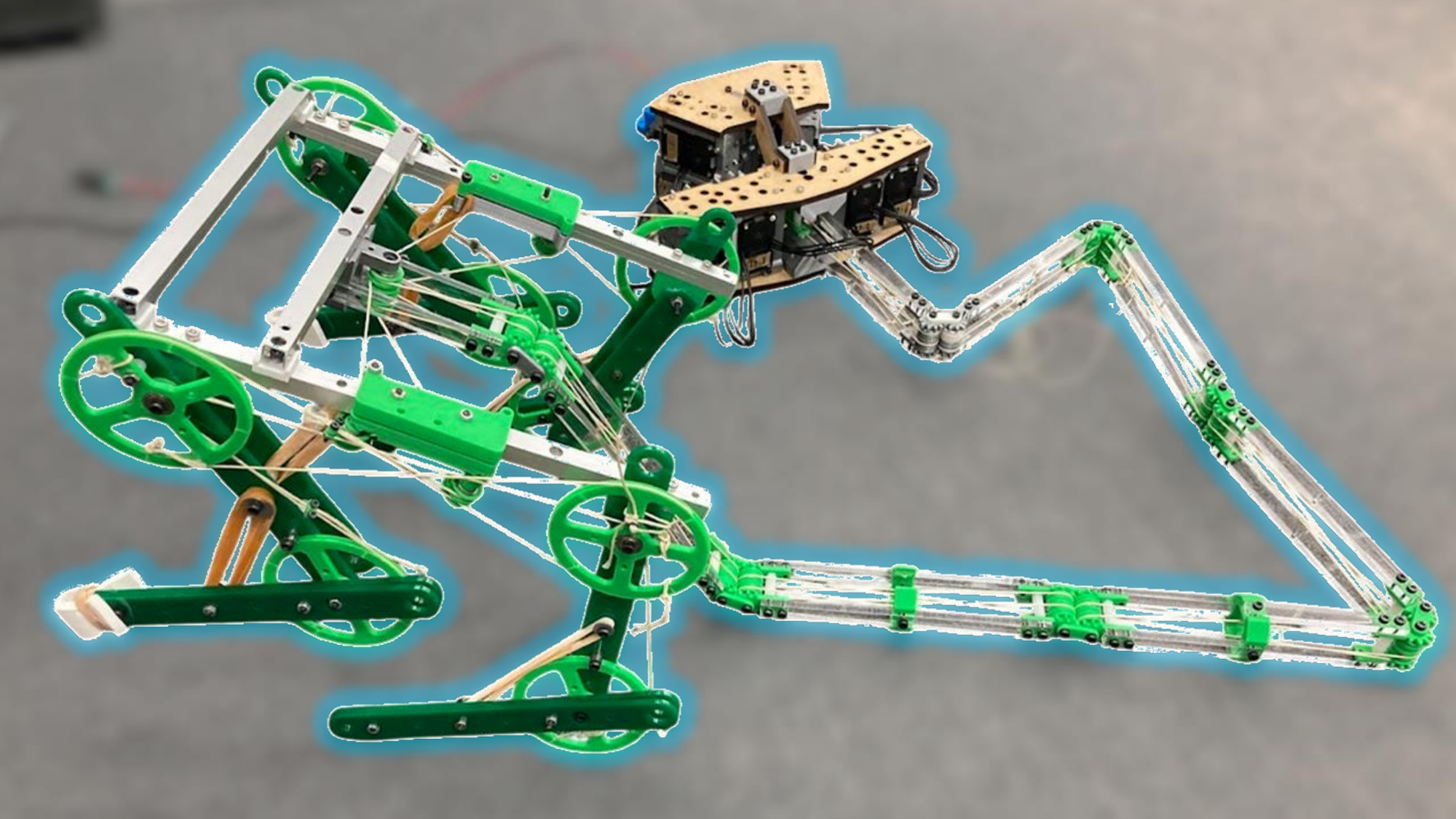}
  \caption{Overview of the remote wire-driven robot we created in this study. The wire wound around the motor concentrated in the brown motor unit drives the green quadruped robot through the translucent Remote Wire Drive. The Remote Wire Drive can passively bend while transmitting power, allowing the quadruped robot to be a freely moving floating link system.}
  \label{fig:fig1}
\end{figure}

Wire-driven robotic arms represent another category of systems capable of long-distance power transmission. For instance, the Super Dragon robotic arm developed by Endo et al. \cite{super_dragon} uses wires to actuate a 10-meter-long arm. Wire-driven systems offer several advantages, including a wide operational temperature range, no requirement for sealed components, and a simple transmission mechanism using only spooling and passive pulleys. The wire-jamming-based robotic hand developed by Tadakuma et al. \cite{wirejamming} demonstrates high fire resistance. In this study, we used a synthetic fiber wire called Vectran \cite{vectran} \cite{vectran_about}, which remains functional in a wide range of temperatures from below -70°C to over 400°C. However, compared to mobile robots, wire-driven robotic arms have limited reachable areas and reduced obstacle avoidance capabilities.

Therefore, we propose a wire-driven remote actuation system, called Remote Wire Drive, which has the potential to be applied in a wider range of environments compared to traditional remotely actuated robots. We designed and controlled both the remote drive unit and the mobile robot, and validated whether the remote-wire-driven system operates properly. An overview of the entire remotely actuated robot system developed in this study is shown in \figref{fig:fig1}.

\section{Method}
\begin{figure*}[t]
  \centering
  \includegraphics[width=1.0\linewidth]{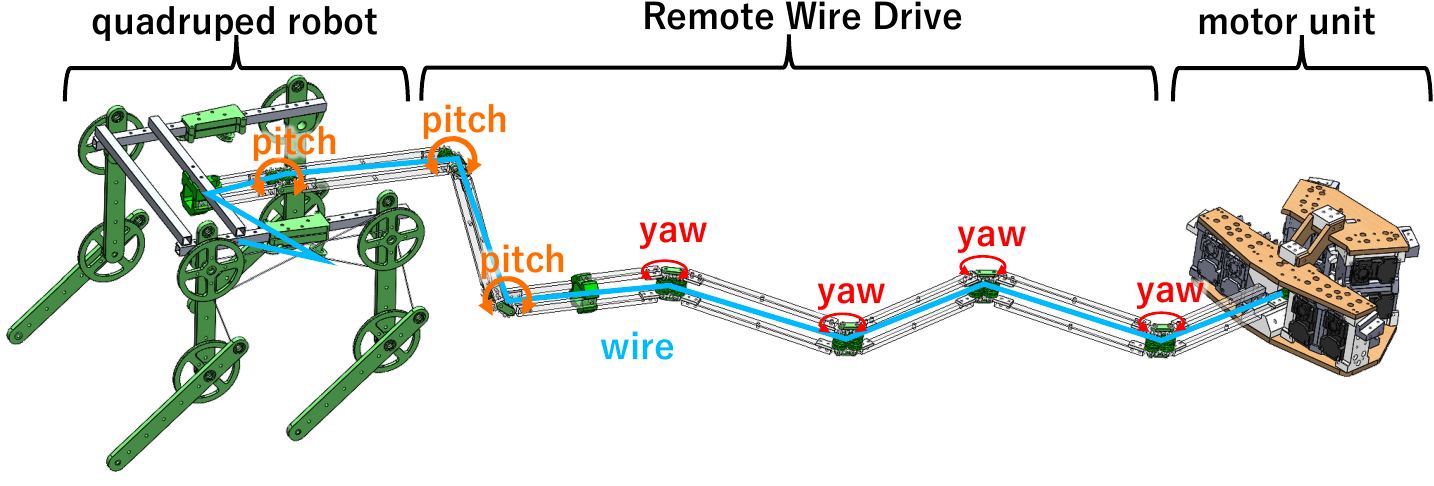}
  \caption{Overall structure of the robot system created in this study. The wire wound around the motor concentrated in the motor unit drives the quadruped robot through a Remote Wire Drive.}
  \label{fig:robot_overview}
\end{figure*}

\figref{fig:robot_overview} illustrates the entire structure of the robot system developed in this study. Wires wound around motors in the motor unit are routed through the Remote Wire Drive and wrap around passive pulleys on the quadruped robot, thus driving it. The specifications are listed in \tabref{tab:specifications}.

\begin{table}[bp]
  \centering
  \begin{tabular}{|l|p{3cm}|}
    \hline
    \textbf{Items} & \textbf{Value} \\
    \hline
    motor model & DYNAMIXEL XM430-W210-R \\ \hline
    motor interface board & DXHUB \\ \hline
    motor number & 8 (each 2 motors are linked) \\ \hline
    wire number & 4 \\ \hline
    maximum wire tension & 200N \\ \hline
    wire material & vectran \\ \hline
    Remote Wire Drive DOF & 7 \\ \hline
    Remote Wire Drive link length & 174mm \\ \hline
    Remote Wire Drive link number & 7 \\ \hline
    quadruped robot DOF & 8 (each 2 joints are coupled) \\ \hline
    leg link length & 125mm \\ \hline
  \end{tabular}
  \caption{Specifications}
  \label{tab:specifications}
\end{table}

\subsection{Design of Remote Wire Drive}
\begin{figure}[tbp]
  \centering
  \includegraphics[width=1.0\linewidth]{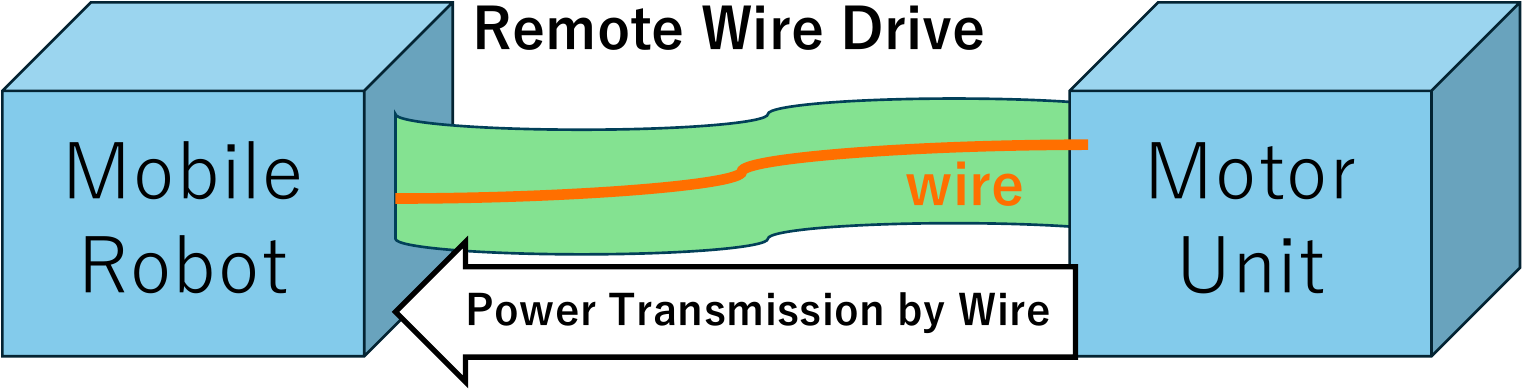}
  \caption{Concept of Remote Wire Drive. The wire wound around the motor concentrated in the motor unit drives the mobile robot through a Remote Wire Drive, which works like a mechanical power cable.}
  \label{fig:rwd_concept}
\end{figure}

We call the general mechanism through which wires are routed to remotely actuate a robot, as shown in \figref{fig:rwd_concept}, as "Remote Wire Drive".

\begin{figure}[tbp]
  \centering
  \includegraphics[width=1.0\linewidth]{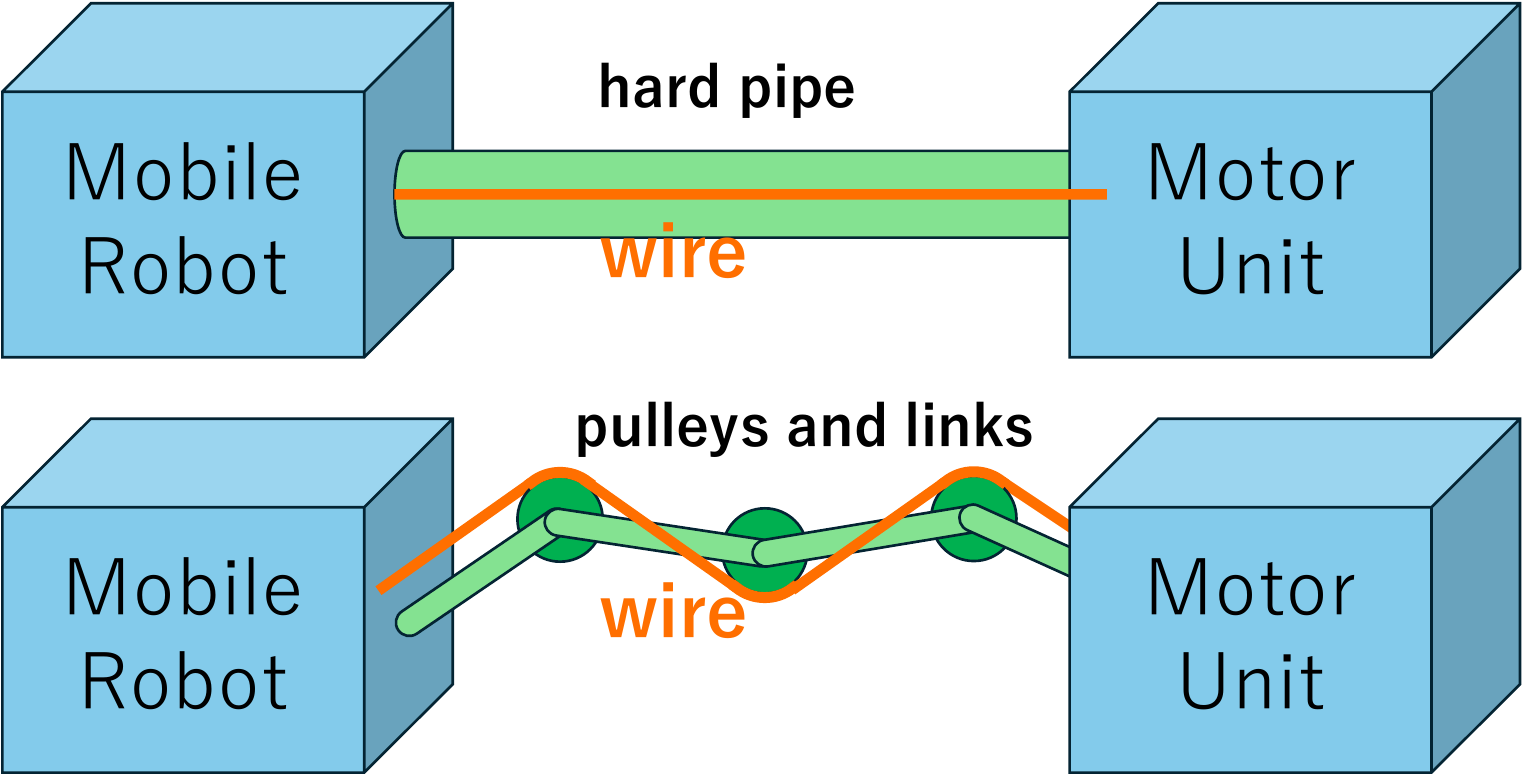}
  \caption{Examples of mechanisms which cannot be used as Remote Wire Drive. The hard pipe hinders the movement of the mobile robot, while the normal pulleys and links act as a robot arm and interfere with the movement of the mobile robot.}
  \label{fig:rwd_nonsuitable}
\end{figure}

To construct a Remote Wire Drive, the transmission mechanism must deform independently of the displacement of the internal wire to avoid interfering with the movement of the mobile robot at the end. For instance, simply running a wire through a hard pipe as in \figref{fig:rwd_nonsuitable} makes it impossible for the robot to move around. Furthermore, using a simple serial link with a single pulley at each joint causes mutual coupling between link motion and wire displacements in \cite{Coupled_tendon_driven}, effectively turning the system into a robotic arm and undermining the advantages of a mobile robot.

One candidate mechanism that deforms independently of internal wire displacement is the Tendon Sheath Mechanism (TSM) \cite{various_tendon_sheath_robots}. Widely used in applications from bicycle brakes to surgical instruments \cite{tsm_endoscopic} and robots \cite{tendon_sheath_robot,tendon_sheath_actuated_manipulator}, TSMs feature the property that sheath bending does not interfere with tendon push-pull motion. Therefore, attaching a mobile robot to the end of a TSM enables it to receive power remotely while maintaining free movement.
However, the Remote Wire Drive used in this study does not adopt the TSM structure. This is because TSM suffers from unstable transmission efficiency due to its efficiency degrading exponentially with cumulative bending angle \cite{bowden_friction}.

\begin{figure}[tbp]
  \centering
  \includegraphics[width=1.0\linewidth]{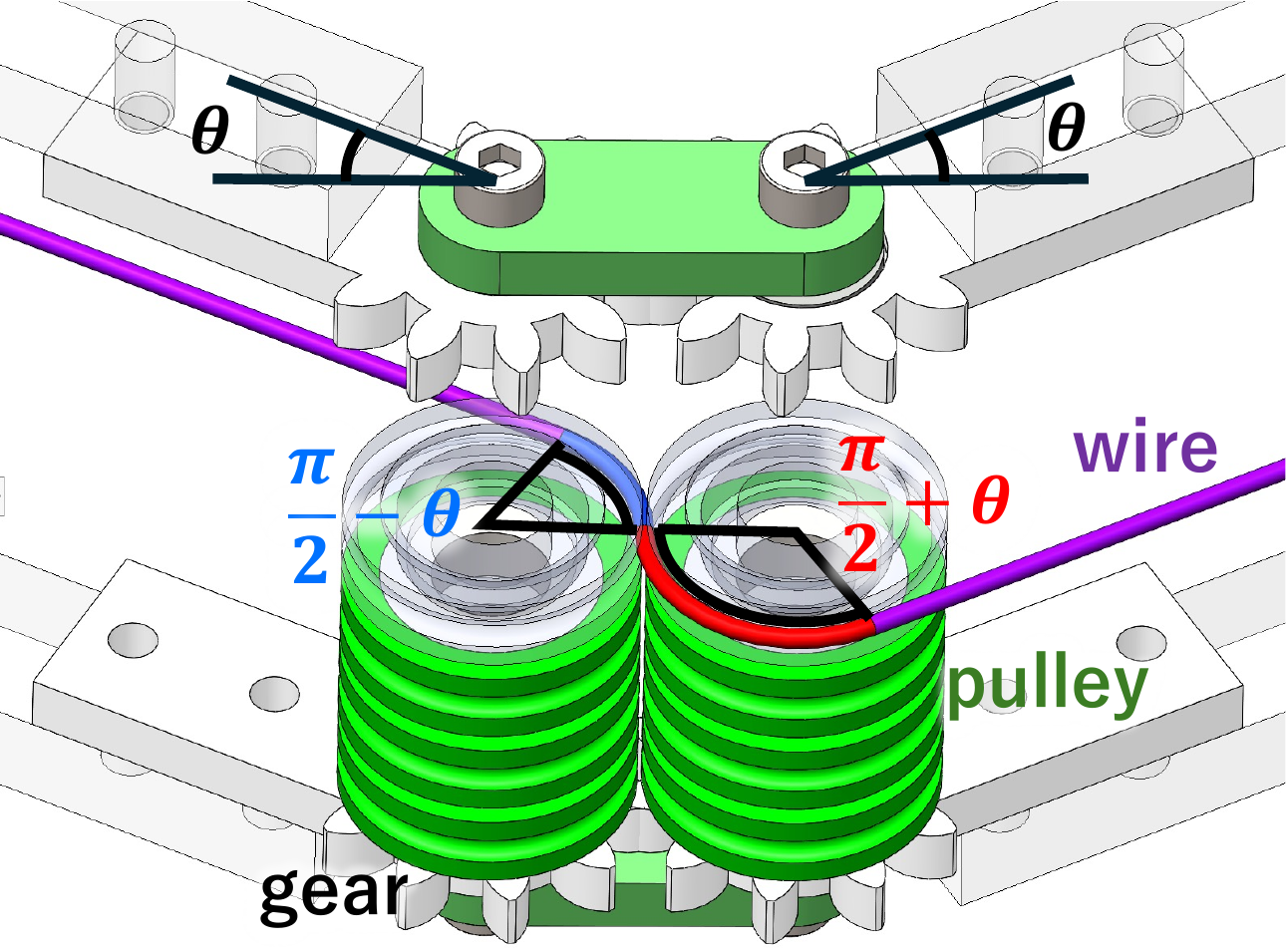}
  \caption{Structure of decoupled joint. When the joint as a whole rotates by $2\theta$, the sum of the wrap angles remains constant at $\pi$. Therefore, the joint displacement and the wire displacement become independent, and according to the principle of work, the joint torque and the wire tension also become independent.}
  \label{fig:decoupled_joint}
\end{figure}

\begin{figure}[tbp]
  \centering
  \includegraphics[width=1.0\linewidth]{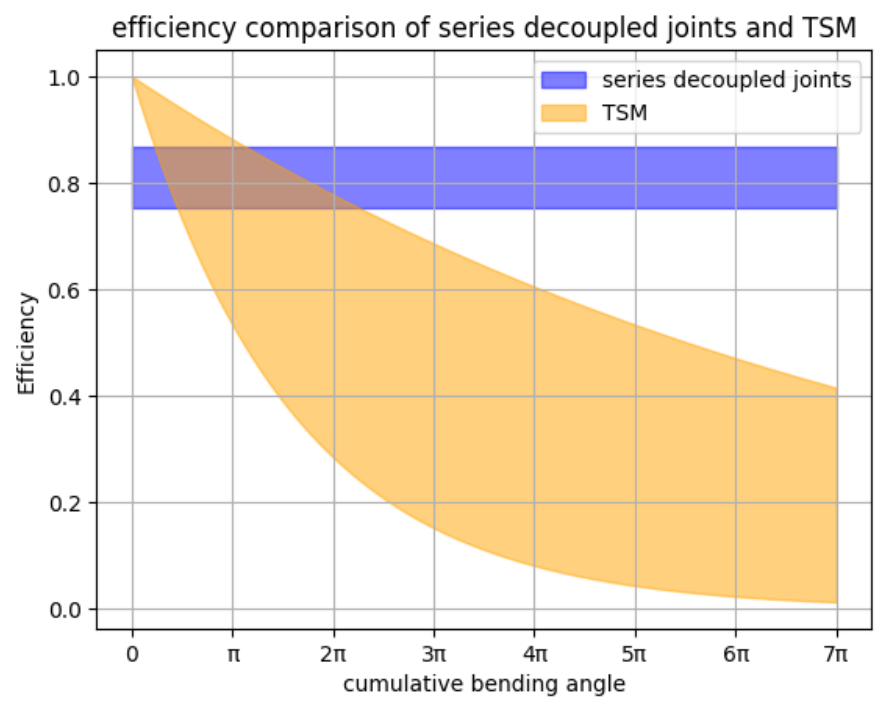}
  \caption{Efficiency ranges of series decoupled joints and TSM as a function of cumulative bending angle. While the TSM exhibits more than a twofold variation in efficiency, the decoupled joint maintains a constant efficiency regardless of the cumulative bending angle and achieves higher efficiency than the TSM across most of the range.}
  \label{fig:efficiency_graph}
\end{figure}

To address this drawback, our Remote Wire Drive uses serially connected \textbf{decoupled joints}. A decoupled joint, as shown in \figref{fig:decoupled_joint}, is a pair of synchronized antagonistic joints that allows the wire displacement to remain independent of joint angle displacement. Though mainly used in robotic arms like D3-ARM \cite{D3_arm} and LIMS \cite{LIMS2} to simplify control, we repurpose them by serially connecting multiple decoupled joints to create a Remote Wire Drive. This achieves deformation independence similar to TSMs, enabling a flexible yet efficient transmission structure. Compared with TSM, our approach offers the following advantages:

\begin{itemize}
  \item  As shown in \figref{fig:decoupled_joint}, each joint contains exactly two pulleys, and the total winding angle over the pulleys remains constant regardless of configuration. This results in consistent friction characteristics.
  \item Unlike TSM, which involves sliding contact, our design exclusively uses bearings with pulleys, allowing high transmission efficiency even through many relay points. In fact, according to the study by Nagai et al. \cite{fiber_rope_basic}, the transmission efficiency of a wire-pulley system depends on the ratio of the pulley diameter D to the wire diameter d, and when $\frac{D}{d}=15$ as in this study, the efficiency ranges from 98\% to 99\% across various wire materials. When the Remote Wire Drive transmits force through 14 pulleys, the overall efficiency is estimated to be between 75\% and 87\%. On the other hand, the transmission efficiency in systems involving friction can be approximated by $e^{-\mu \theta}$, where $\mu$ is the coefficient of friction and $\theta$ is the cumulative wrap angle. When using PTFE, a material known for its low friction in sheaths, the coefficient of friction  $\mu$ is estimated to range from approximately 0.04 to 0.2. Based on these considerations, the efficiency ranges of the TSM and the serially connected decoupled joints are compared as shown in \figref{fig:efficiency_graph}.
\end{itemize}

\begin{figure}[tbp]
  \centering
  \includegraphics[width=1.0\linewidth]{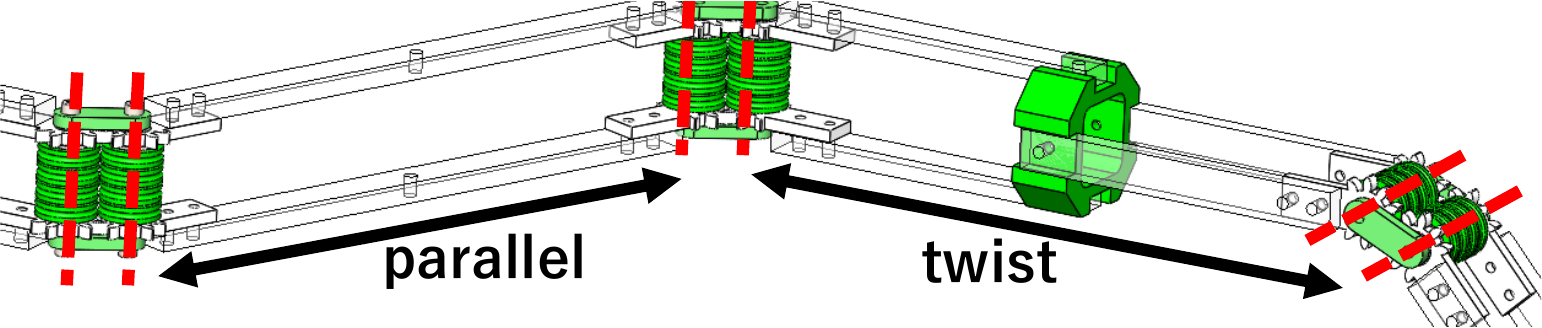}
  \caption{Link types used for serial decoupled joints. One has parallel joint placement and the other has twisted joint placement. This allows for a three-dimensional range of movement, not just on a plane.}
  \label{fig:dj_links}
\end{figure}

The serial connection of decoupled joints utilizes two types of links, as shown in \figref{fig:dj_links}. By combining these appropriately, the Remote Wire Drive can accommodate both yaw and pitch degrees of freedom, as illustrated in \figref{fig:robot_overview}.

\subsection{Design and Control of the Quadruped Mobile Robot}

This section describes the design and control of the mobile robot at the end of the system. To verify the capability of remotely actuating the mobile robot via the Remote Wire Drive, it was necessary to develop a wire-driven mobile robot. Although the simplest form of mobility is wheels, constructing mechanisms with continuous rotation (such as wheels) is difficult with wire drive. Therefore, a legged mechanism was adopted.

\begin{figure}[tbp]
  \centering
  \includegraphics[width=1.0\linewidth]{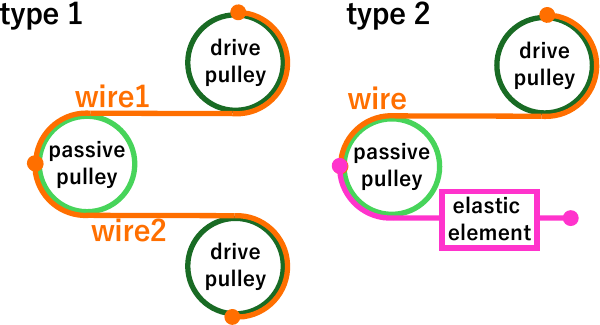}
  \caption{Classification of drive methods in wire-driven systems. In one, two wires compete with each other, and in the other, the single wire competes with a passive elastic element.}
  \label{fig:drive_types}
\end{figure}

In a wire-driven robot, where the drive pulley connected to a motor transmits force through wires to a passive pulley, the drive method can be broadly classified into two types, as shown in \figref{fig:drive_types}:

\begin{itemize}
  \item Type 1: Two wires connected to separate drive pulleys antagonistically drive a single passive pulley.
  \item Type 2: One wire connected to a drive pulley and another connected to an elastic element antagonistically drive a single passive pulley.
\end{itemize}

To minimize the width of the Remote Wire Drive, we adopted Type 2.

\subsubsection{Leg Design and Control}
To provide turning capability, a symmetrical propulsion system was employed. Although we wanted to minimize the number of wires, it was difficult to construct a propulsion mechanism with only one wire per side. Thus, we opted for two wires per side. When designing a legged locomotion mechanism using two antagonistic wire pairs per leg in combination with elastic elements, the simplest configuration is a bipedal system. However, bipedal standing requires sensors such as IMUs and a high temporal response, which would conflict with our study’s concept of avoiding electronics. Moreover, given that power is transmitted through wires approximately 1.2 meters in length, achieving a high temporal response is also challenging. Therefore, we implemented a quadruped configuration with front and back legs coupled.

\begin{figure}[tbp]
  \centering
  \includegraphics[width=1.0\linewidth]{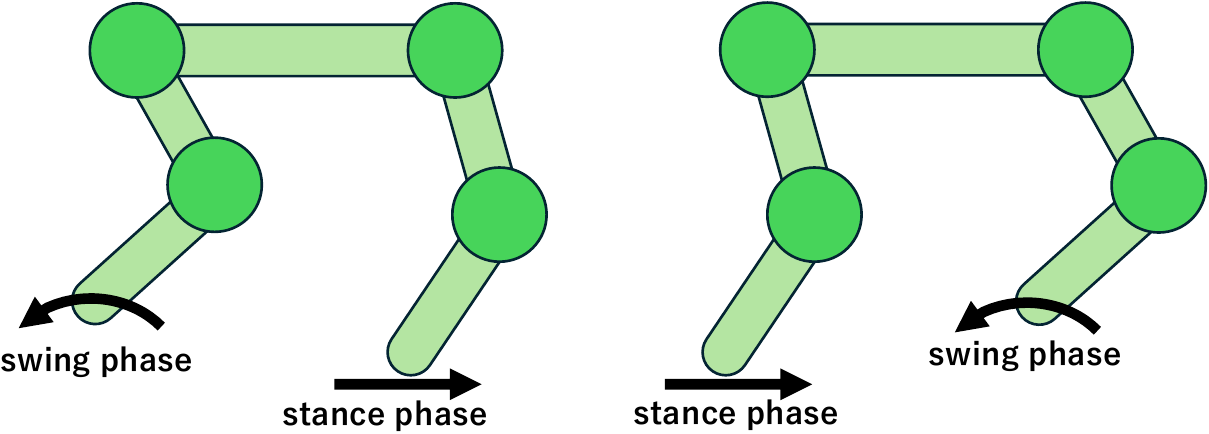}
  \caption{Cycle of the trot gait. The front and back legs alternate between swing and stance.}
  \label{fig:trot}
\end{figure}

Various gait patterns exist for quadrupeds \cite{gait_patterns}, but we modeled a trot gait where front and back legs are synchronized, as shown in \figref{fig:trot}.

\begin{figure}[tbp]
  \centering
  \includegraphics[width=1.0\linewidth]{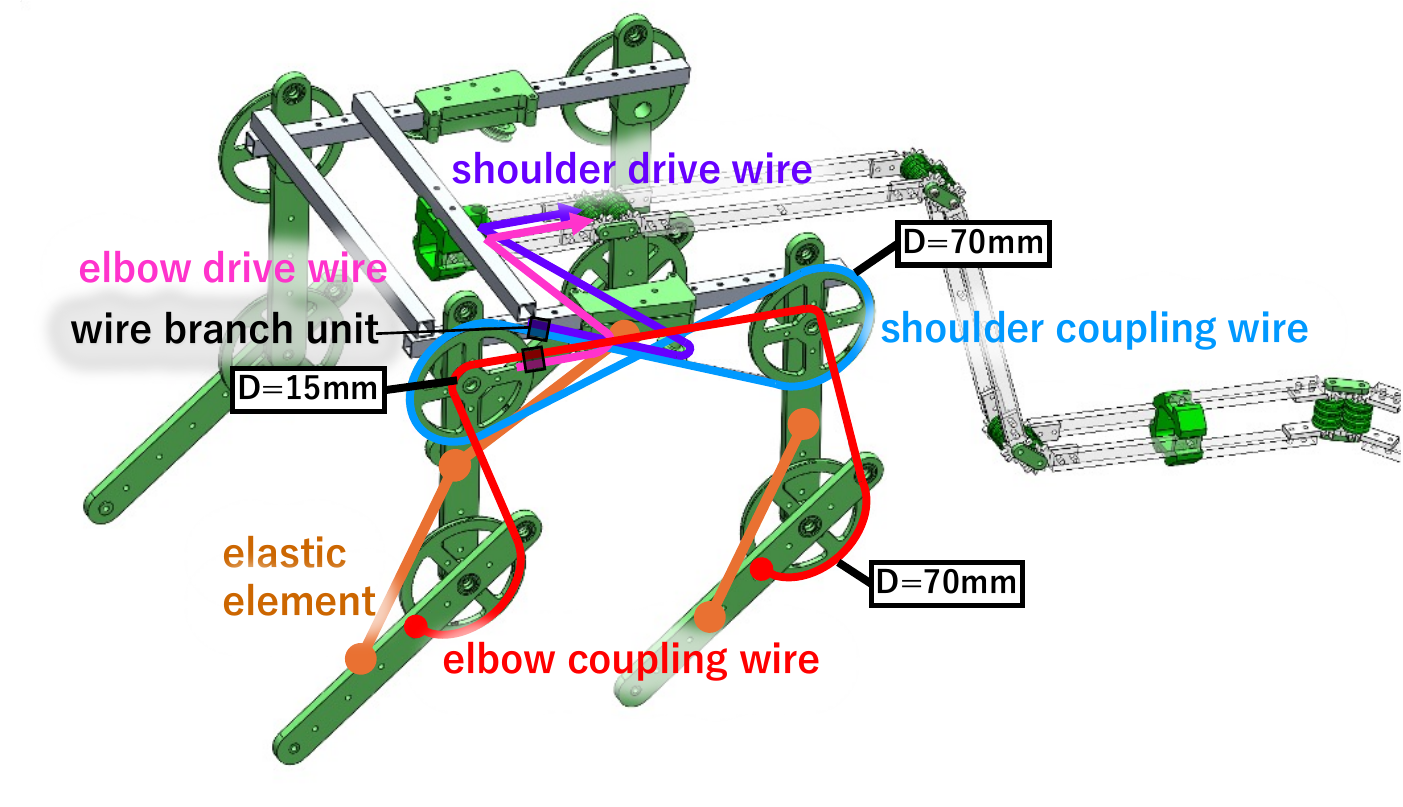}
  \caption{Wire placement of the robot. The front and back shoulders and elbows are coupled in opposite directions.}
  \label{fig:wire_placement}
\end{figure}

To implement this gait, we devised the wire layout shown in \figref{fig:wire_placement}. The shoulder and elbow joints are coupled in opposite directions between the front and back legs. The coupling wire is driven in conjunction with the drive wire via a wire branch unit. The antagonistic elastic element maintains constant wire tension.

We define the trajectory of the legs mathematically. In a wire-driven system, given the wire length vector relative to the initial state $\bm{l}$, joint angle vector $\bm{q}$, and initial joint angle vector $\bm{q}_0$, the following equation holds:

\begin{equation}
  \bm{l} = \bm{G} (\bm{q} - \bm{q}_0)
  \label{eq:tendon_basic_eq}
\end{equation}

Here, $\bm{G}$ is referred to as the tendon Jacobian. In our robot, coupling wire displacement $\bm{l}$ is identical for the front and back legs. Defining the joint angle vectors, initial joint angles, and tendon Jacobians as in \tabref{tab:symbol_definition}, we obtain:

\begin{equation}
  \bm{l} = \bm{G}^f (\bm{q}^f - \bm{q}^f_0) = \bm{G}^b (\bm{q}^b - \bm{q}^b_0)
  \label{eq:coupled_eq}
\end{equation}

\begin{table}[tbp]
  \begin{tabular}{|l|l|l|}
    \hline
    & front leg & back leg \\
    \hline
    joint angle vector & $\bm{q}^f$ & $\bm{q}^b$ \\ \hline
    initial joint angle vector & $\bm{q}^f_0$ & $\bm{q}^b_0$ \\ \hline
    tendon Jacobian & $\bm{G}^f$ & $\bm{G}^b$ \\
    \hline
  \end{tabular}
  \caption{Symbol definitions}
  \label{tab:symbol_definition}
\end{table}

\begin{figure}[tbp]
  \centering
  \includegraphics[width=1.0\linewidth]{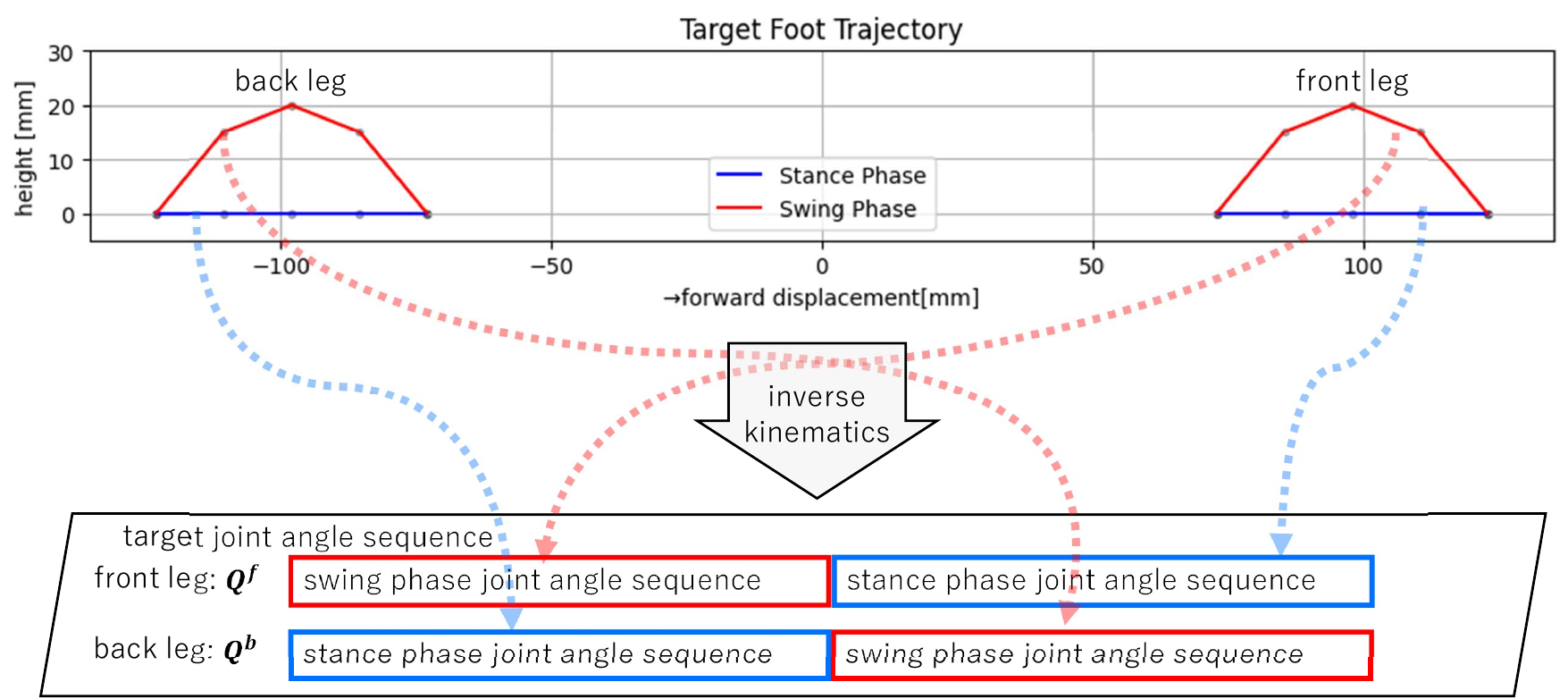}
  \caption{Target joint sequence calculated from the target foot trajectory. Each sequence has a length of N for the swing and stance phases.}
  \label{fig:target_joint_seq}
\end{figure}

Based on this, we performed optimization to determine the design parameters $\bm{G}^f$, $\bm{G}^b$, $\bm{q}^f_0$, and $\bm{q}^b_0$. The ideal foot trajectory in a trot gait is a comb-like shape, as shown in \figref{fig:target_joint_seq}. Here, the ground plane is defined to be 200 mm below the body. To approximate this ideal as closely as possible, we formulated the optimization problem in \equref{eq:optim} . The constraints are imposed not only due to coupling requirements but also for design simplicity, such as reducing the number, size, and types of the pulleys. The ideal time-series joint angle vector sequences $\bm{Q}^f$ and $\bm{Q}^b$ are derived via inverse kinematics (IK) to follow the top trajectory in \figref{fig:target_joint_seq} . As shown in \figref{fig:target_joint_seq} , swing and stance phases are complementary in $\bm{Q}^f$ and $\bm{Q}^b$ .

\begin{equation}
    \begin{aligned}
    \min_{\bm{G}^f, \bm{G}^b, \bm{q}_0^f, \bm{q}_0^b} 
    \sum_{i=1}^{2N} \left| 
    \bm{G}^f\left(\bm{Q}^f[i] - \bm{q}_0^f\right) - \bm{G}^b\left(\bm{Q}^b[i] - \bm{q}_0^b\right) 
    \right|^2 \\
    \text{s.t.} \quad 
    \begin{cases}
    \bm{G}^f\left(\bm{q}^f - \bm{q}_0^f\right) = \bm{G}^b\left(\bm{q}^b - \bm{q}_0^b\right) \\
    \|\bm{G}\|_2 > 60 \quad (\bm{G} = \bm{G}^f, \bm{G}^b) \\
    \|\bm{G}\|_\infty < 40 \quad (\bm{G} = \bm{G}^f, \bm{G}^b) \\
    \bm{G}(1,2) = 0 \quad (\bm{G} = \bm{G}^f, \bm{G}^b) \\
    |\bm{G}(1,1)| = |\bm{G}(2,2)| \quad (\bm{G} = \bm{G}^f, \bm{G}^b) \\
    |\bm{G}^f(0,0)| = |\bm{G}^b(0,0)|
    \end{cases}
  \end{aligned}
  \label{eq:optim}
\end{equation}

\begin{figure}[tbp]
  \centering
  \includegraphics[width=1.0\linewidth]{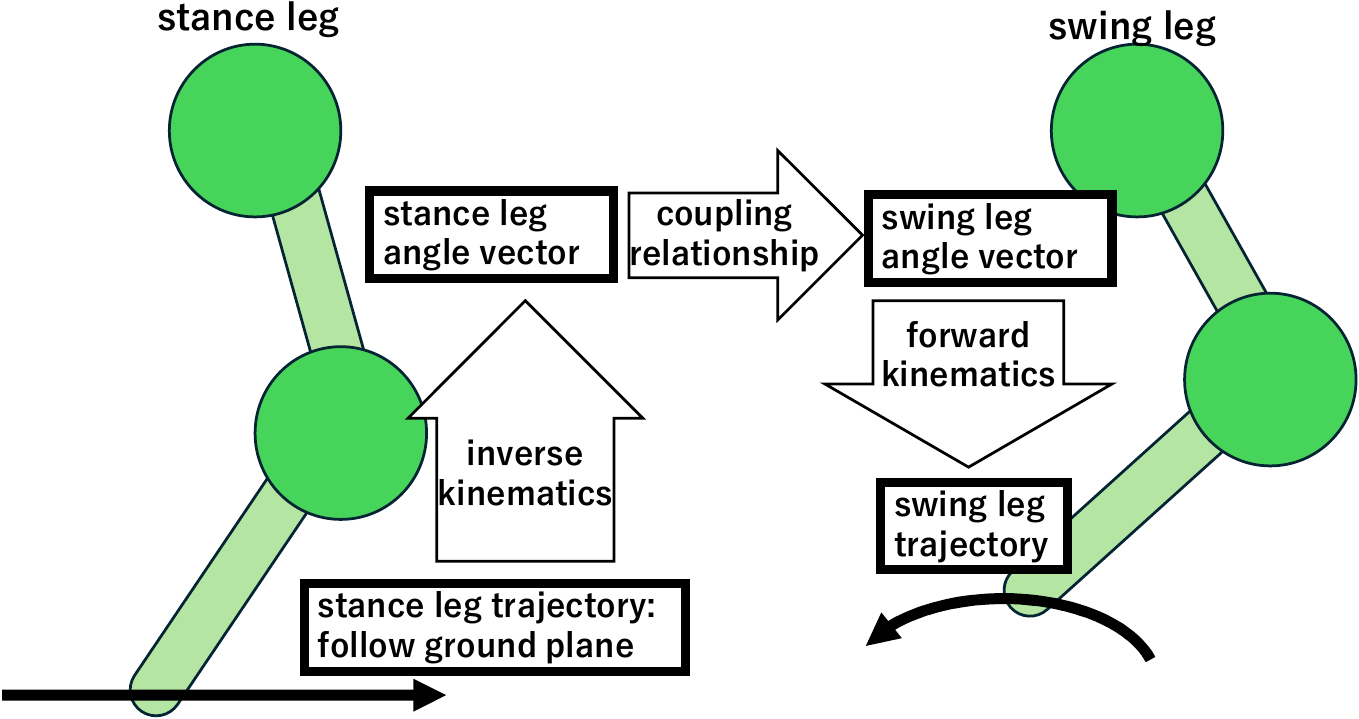}
  \caption{Method of calculating the swing leg trajectory. The stance leg angle vector is computed from the stance leg trajectory using inverse kinematics. Subsequently, the swing leg angle vector is derived from this using \equref{eq:coupled_eq} , and the swing leg trajectory is obtained through forward kinematics.}
  \label{fig:swing_calc}
\end{figure}

\begin{figure}[tbp]
  \centering
  \includegraphics[width=1.0\linewidth]{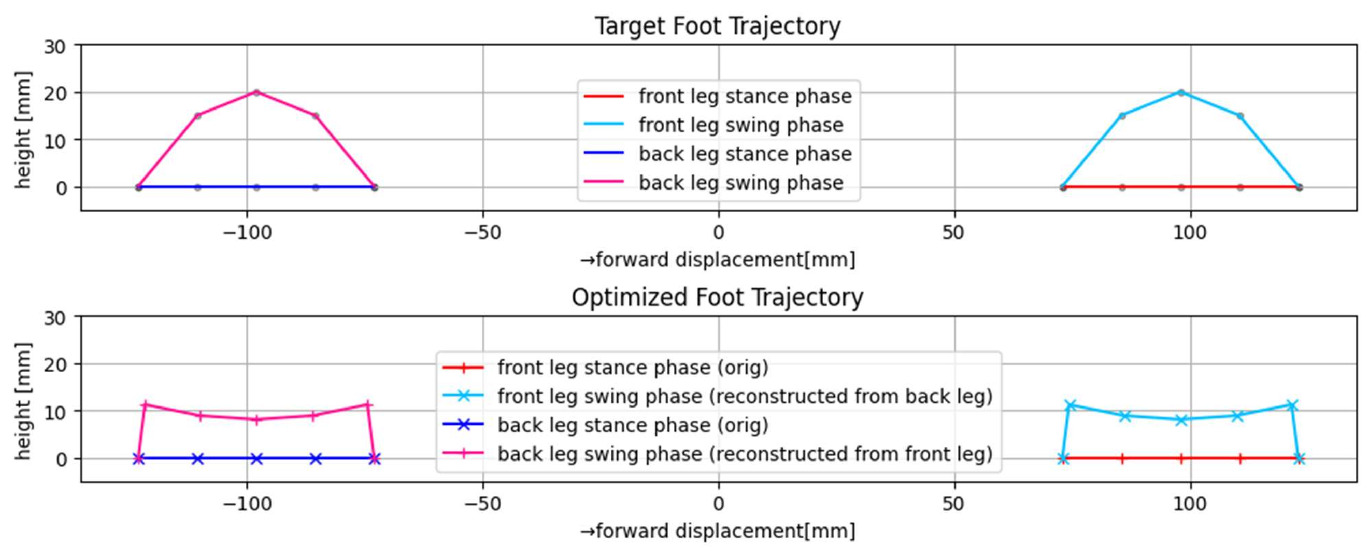}
  \caption{Target foot trajectory and optimized foot trajectory. We sought a leg trajectory closest to the ideal target trajectory within the constraints of the leg coupling.}
  \label{fig:traj_optimize}
\end{figure}

The results are shown in \equref{eq:result_eq}. Using these design parameters, we made the stance leg follow the ground plane, and calculated the swing leg trajectory using the method described in \figref{fig:swing_calc}. The resulting trajectory is shown in \figref{fig:traj_optimize}.

\begin{equation}
  \begin{aligned}
  \bm{G}^f = 
  \begin{pmatrix}
  40 & 0 \\
  20 & -40
  \end{pmatrix}, \quad
  \bm{G}^b = 
  \begin{pmatrix}
  -40 & 0 \\
  -20 & 40
  \end{pmatrix} \\
  \bm{q}_0^f = 
  \begin{pmatrix}
  2.2 \\
  1.9
  \end{pmatrix}, \quad
  \bm{q}_0^b = 
  \begin{pmatrix}
  2.2 \\
  1.7
  \end{pmatrix}
  \end{aligned}
  \label{eq:result_eq}
\end{equation}

These values were then fine-tuned, taking into account design simplicity and other practical considerations, and finalized as follows.

\begin{equation}
  \begin{aligned}
  \bm{G}^f = 
  \begin{pmatrix}
  35 & 0 \\
  7.5 & -35
  \end{pmatrix}, \quad
  \bm{G}^b = 
  \begin{pmatrix}
  -35 & 0 \\
  7.5 & 35
  \end{pmatrix} \\
  \bm{q}_0^f = 
  \begin{pmatrix}
  2.6 \\
  1.0
  \end{pmatrix}, \quad
  \bm{q}_0^b = 
  \begin{pmatrix}
  1.7 \\
  2.6
  \end{pmatrix}
  \end{aligned}
  \label{eq:actual_value}
\end{equation}

\begin{figure}[tbp]
  \centering
  \includegraphics[width=1.0\linewidth]{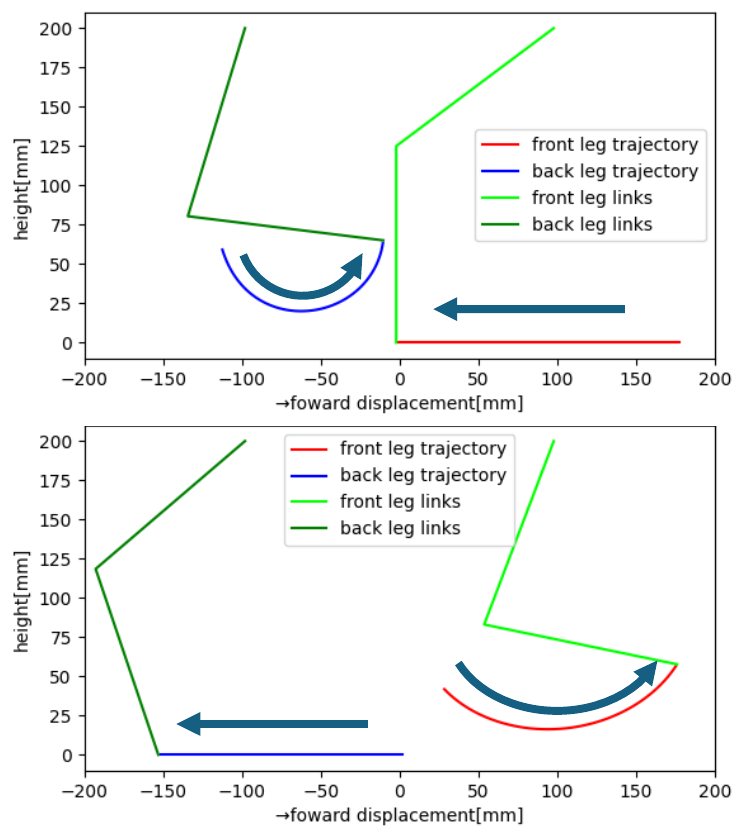}
  \caption{Foot trajectory plan. The stance trajectory follows the ground plane, and the swing leg trajectory is calculated accordingly based on the coupling with the stance leg joint.}
  \label{fig:links_and_trajectory}
\end{figure}

Using inverse kinematics (IK), the stance leg was made to follow a ground plane 200 mm below the body, as shown in \figref{fig:trot}. Applying the trajectory from the stance leg to the swing leg using the method in \figref{fig:swing_calc} resulted in a trajectory where the swing leg remains above the ground, regardless of whether the front or back leg is in stance, as shown in \figref{fig:links_and_trajectory}.

\begin{figure}[tbp]
  \centering
  \includegraphics[width=1.0\linewidth]{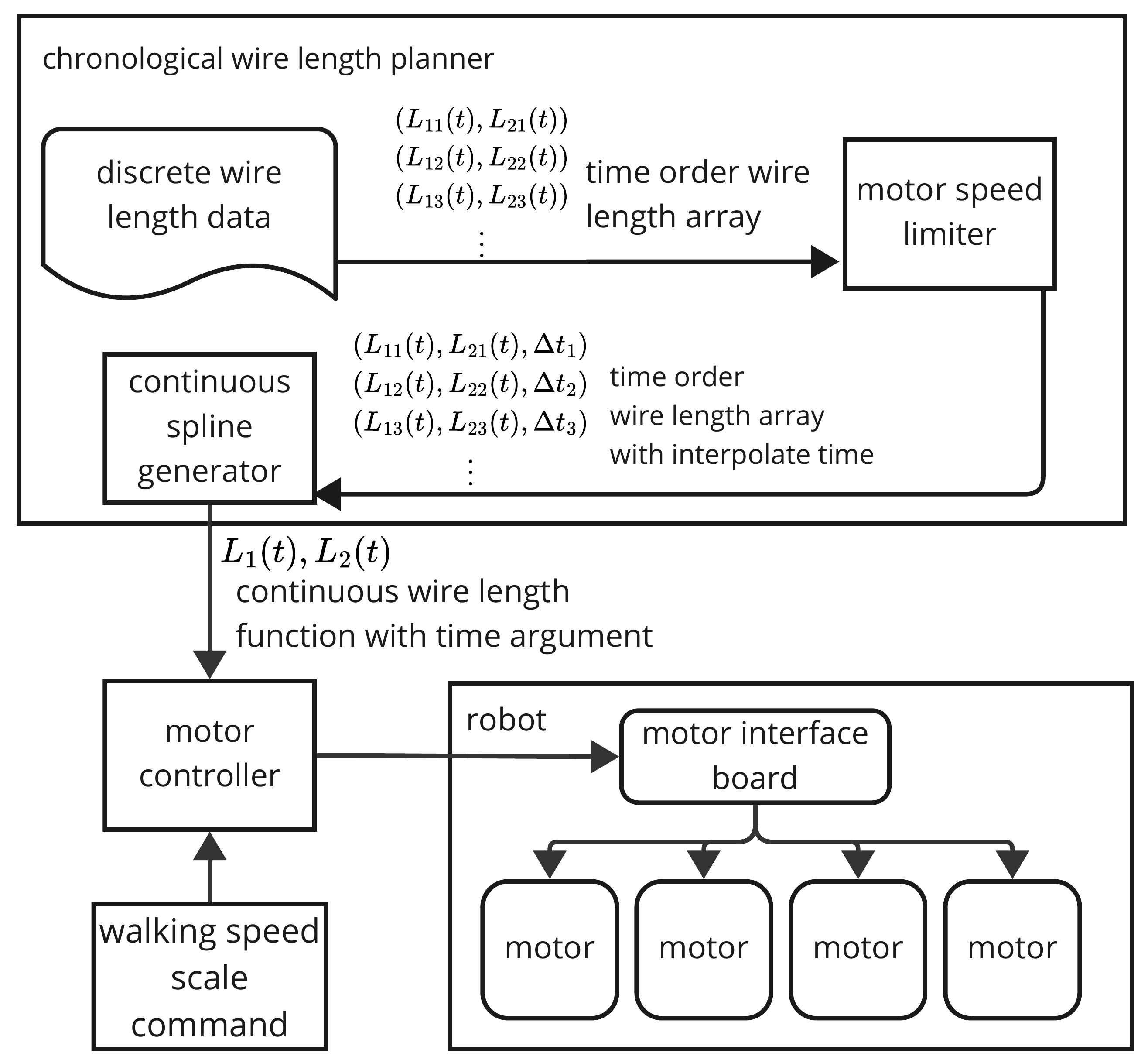}
  \caption{System of the robot. Based on the time-series wire length sequence, an interpolation time is computed to ensure motor velocity stays below a threshold. A continuous spline function is then generated from this and queried by the motor controller, scaled in time by a walking-scale command before being sent to the motors.}
  \label{fig:system}
\end{figure}

To control the physical robot, we concatenate the joint trajectories for when each of the front and back legs are in stance. Then we calculate the time-series of the drive wire displacement $\bm{l}$, which is mechanically synchronized to the coupling wire displacement, using \equref{eq:coupled_eq} .Next, as shown in \figref{fig:system}, time interpolation is added to the time-series data to ensure that the motor speed remains below a certain threshold. The data is then made continuous through spline interpolation and time-scaled according to the walking scale command before being sent as motor commands.

\section{Experiment}
\subsection{Walking Motion Midair}
\begin{figure}[tbp]
  \centering
  \includegraphics[width=1.0\linewidth]{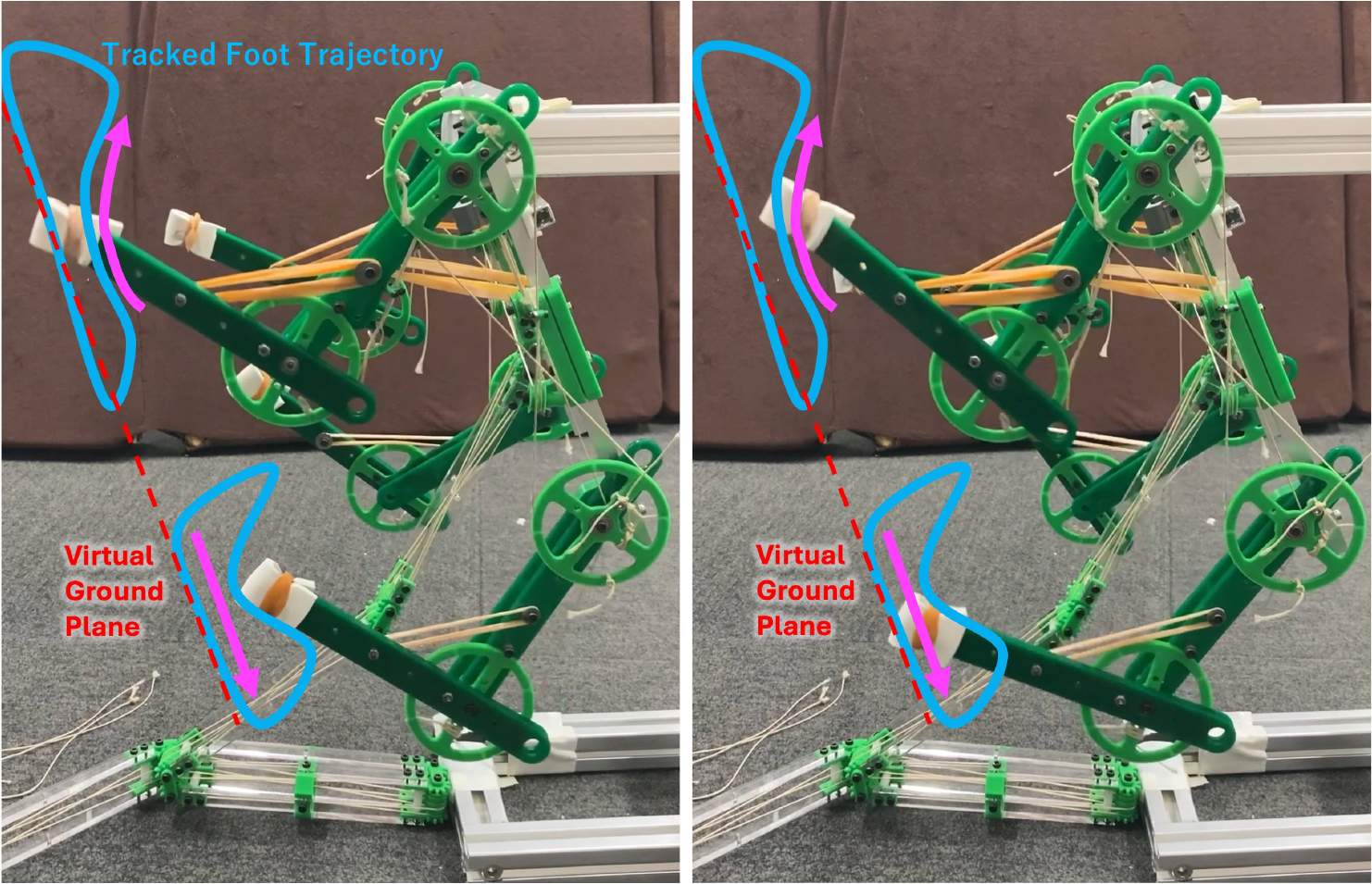}
  \caption{Quadruped robot performing walking motion midair. Tracked trajectories shown in light blue lines.}
  \label{fig:airwalk}
\end{figure}

\begin{figure}[tbp]
  \centering
  \includegraphics[width=1.0\linewidth]{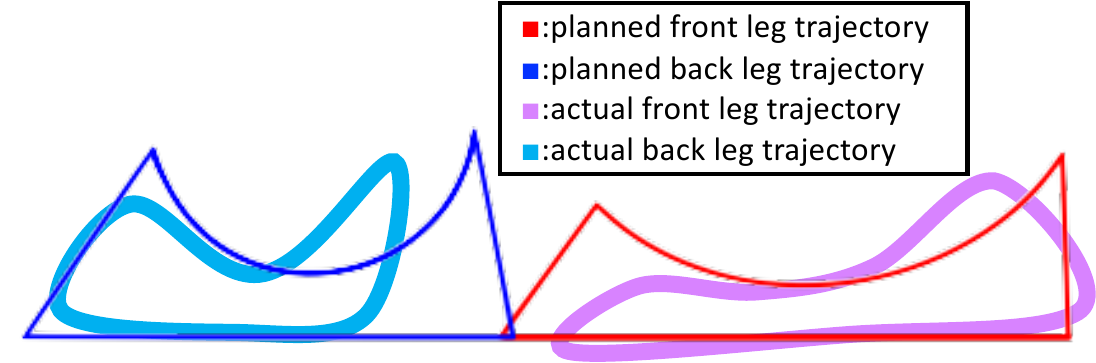}
  \caption{Comparison of planned and actual trajectories. They are mostly consistent.}
  \label{fig:trajectory_comparison}
\end{figure}

We fixed the suspended robot to an aluminum frame and executed the walking trajectory midair to verify leg trajectory. As shown in \figref{fig:airwalk}, the robot successfully alternated between swing and stance phases. Furthermore, the stance trajectories of the front and back legs followed the intended virtual ground plane. As shown in \figref{fig:trajectory_comparison}, the commanded and actual trajectories generally agree; however, some deviations are observed, primarily in the back leg.

\subsection{Walking on Floor}
\begin{figure*}[!t]
  \centering
  \includegraphics[width=1.0\linewidth]{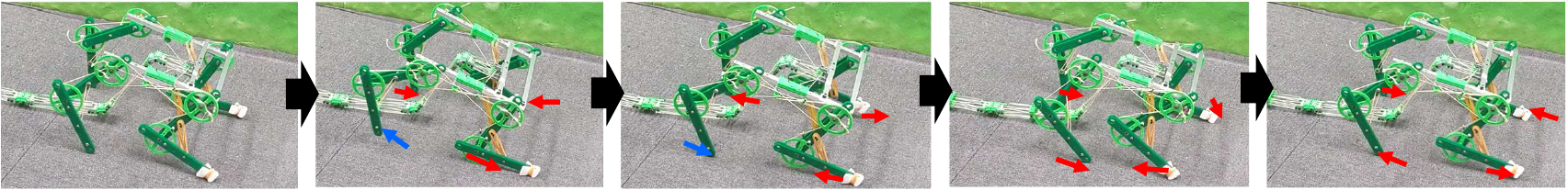}
  \caption{Quadruped robot walking on floor. Arrows indicate foot directions. Red arrows indicate the contact between the leg and the floor. Blue arrows indicate that the leg is above the floor.}
  \label{fig:time_gait_sequence}
\end{figure*}

\begin{figure}[tbp]
  \centering
  \includegraphics[width=1.0\linewidth]{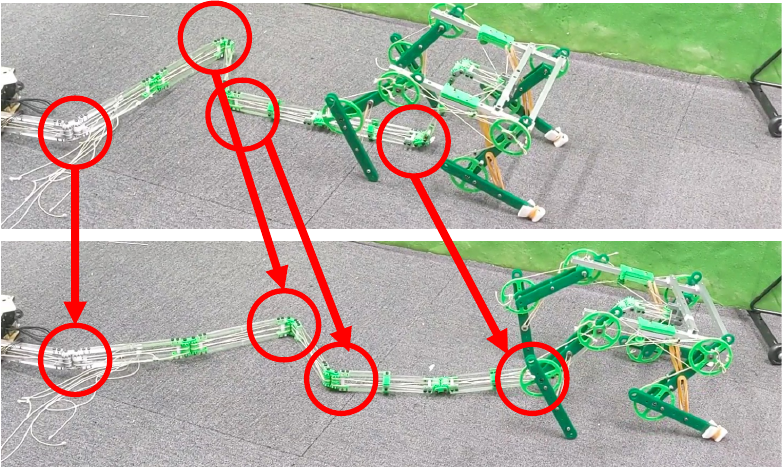}
  \caption{Change in Remote Wire Drive configuration from the beginning to end of walking. Significant variations were observed across 4 decoupled joints.}
  \label{fig:rwd_config_change}
\end{figure}

The actual floor walking experiment is illustrated in \figref{fig:time_gait_sequence}. The robot advanced approximately 30 cm over 30 seconds. During this time, the configuration of the Remote Wire Drive changed across several decoupled joints as shown in \figref{fig:rwd_config_change}. However, no change was observed in leg behavior, confirming that the Remote Wire Drive functioned properly without impeding locomotion.

\section{Discussion}

In the midair walking experiment, significant discrepancies from the planned trajectory were observed. Possible causes include the non-negligible stretch of the 1.2-meter-long wires and initial joint angle errors caused by manually tying the wire endpoints.

In the field of surgical instruments using Tendon-Sheath Mechanisms (TSMs), prior studies have addressed the issue of stretch compensation \cite{tsm_elongation} \cite{tsm_elong_modeling}. The Remote Wire Drive used in this study is theoretically more stable in terms of transmission efficiency compared to TSMs, and its tension distribution is more uniform, making wire elongation modeling more feasible. Further investigation is necessary in future work.

In the walking experiment, we verified that a mobile robot can achieve autonomous locomotion via the Remote Wire Drive. Moreover, despite continuous changes in the configuration of the Remote Wire Drive, the robot was able to repeatedly perform consistent motions. This confirms that a serially connected decoupled joints system can function as an effective remote actuation mechanism.

On the other hand, the quadruped robot dragged its swing legs across the floor while walking, which was contrary to the intended behavior. Since the trot gait involves a phase where the body is supported by only two legs, it is inherently difficult to maintain balance with a simple open-loop gait pattern lacking feedback control. Nevertheless, the robot was able to move forward, indicating that some level of functional differentiation between stance and swing legs was achieved—potentially through implicit load distribution.

\section{Conclusion}

In this study, we proposed a new actuation mechanism for remotely driving mobile robots, named the Remote Wire Drive, and realized it in the form of serially decoupled joints. We also built a quadruped robot actuated by this Remote Wire Drive and demonstrated that the remote actuation system can be practically implemented using serial decoupled joints.

However, the quadruped robot developed in this study exhibited limitations such as slow locomotion speed and low control accuracy. We believe that these issues can be mitigated by developing better software for compensating wire friction and elongation, as well as by designing better wire-driven mobile robots with more legs or higher degrees of freedom. Such improvements may bring the Remote Wire Drive closer to practical application.

{
  \bibliographystyle{IEEEtran}
  \bibliography{main}
}

\end{document}